\documentclass[10pt]{wlscirep}
\usepackage[utf8]{inputenc}
\usepackage[T1]{fontenc}
\usepackage{amsmath,amssymb}
\usepackage{booktabs}
\usepackage{array}
\usepackage[table]{xcolor}
\usepackage{float}
\usepackage{placeins}
\usepackage{pifont}
\newcommand{\comment}[1]{}

\usepackage{multirow}   
\usepackage{colortbl}   
\usepackage{xcolor}     
\usepackage{graphicx} 
\usepackage{array}

\definecolor{headercolor}{RGB}{216, 216, 216} 
\definecolor{rowcolor}{RGB}{242, 242, 242}   
\definecolor{changecolor}{RGB}{180, 0, 0}

\title{FairGen: Preference-Aligned Diffusion for Demographically Equitable Medical Image Synthesis}

\author[1,+]{Zhimin Li}
\author[2,+]{Ruichen Zhang}
\author[3,+]{Zhen Tan}
\author[4]{Howard J Aizenstein}
\author[1,*]{Jingtong Hu}
\author[2,*]{Tianlong Chen}

\affil[1]{University of Pittsburgh, Swanson School of Engineering, Pittsburgh, 15261, USA}
\affil[2]{The University of North Carolina at Chapel Hill, Department of Computer Science, Chapel Hill, 27599, USA}
\affil[3]{Arizona State University, School of Computing and Augmented Intelligence, Tempe, 85281, USA}
\affil[4]{University of Pittsburgh, Department of Psychiatry, Pittsburgh, 15213, USA}

\affil[*]{Corresponding authors: jthu@pitt.edu, tianlong@cs.unc.edu}

\affil[+]{these authors contributed equally to this work}

\begin{abstract}

Medical imaging is central to modern diagnostics, and artificial intelligence (AI) systems are increasingly used to support image-based analysis by improving efficiency, accuracy, and access to care. However, inequities in healthcare access and differential disease prevalence create severe demographic imbalances in clinical image data. Such imbalances are compounded by the fact that diseases can manifest with distinct features across demographic groups, rendering certain phenotypic presentations naturally rare. AI models trained on such imbalanced data risk perpetuating diagnostic bias and widening healthcare disparities. Here we introduce FairGen, a fairness-aware diffusion framework that synthesizes demographically balanced medical images while preserving pathology-relevant visual features. By embedding physician-aligned preferences into the generation process, FairGen improves subgroup coverage during synthesis and downstream classification. Applied to dermatology, radiology, and neuroimaging benchmark tasks, FairGen achieves fairness improvements of 95.9\% for skin images, 80.0\% for chest radiography, and 35.2\% for brain MRI, while maintaining competitive diagnostic accuracy relative to models trained on original clinical data. Clinician-facing expert review and external validation on independent cohorts further support that these gains extend beyond standard fidelity metrics and are not confined to the original in-distribution datasets.

\end{abstract}
\begin{document}

\flushbottom
\maketitle

\thispagestyle{empty}

\section*{Introduction}

Medical imaging is central to modern healthcare, underpinning diagnostics, treatment planning, and medical education across modalities such as computed tomography (CT) \cite{ardila2019end}, magnetic resonance imaging (MRI) \cite{knoll2020advancing}, and dermoscopy \cite{liu2020deep}. Its widespread adoption has transformed clinical practice: in the United States, CT utilization more than doubled between 2000 and 2016 \cite{smith2019trends}, while large-scale resources such as CheXpert, with over 220,000 chest radiographs from more than 65,000 patients, now support research and training \cite{irvin2019chexpert}. Yet the continued reliance on expert interpretation and the scarcity of rare disease cases impose significant limitations. Even in tertiary centers, uncommon presentations such as pediatric brain tumors remain underrepresented, constraining both dataset curation and medical training \cite{lee2024international}. Moreover, the interpretation of medical images requires large-scale investments in expert personnel, imaging infrastructure, and operational resources; for example, millions of annual mammograms place heavy burdens on radiologists and drive up healthcare costs \cite{mckinney2020international}.

Artificial intelligence (AI) offers powerful tools to address these challenges\cite{ma2024segment, wang2025need, aggarwal2021diagnostic}. Diagnostic models trained on large imaging datasets can match or exceed expert performance in specific tasks, as demonstrated by CheXNeXt for chest radiography \cite{rajpurkar2017chexnet}, while AI-based screening approaches have shown economic advantages, such as saving about 140 dollars per patient in pediatric diabetic retinopathy screening compared to provider-based exams \cite{ahmed2025cost}. Generative AI further extends this potential by creating synthetic images to augment scarce datasets and capture rare but clinically important presentations. For example, DermGAN synthesized rare skin cancers across diverse skin tones, enriching datasets with rare disease cases \cite{ghorbani2020dermgan}. More broadly, prior reviews and modality-specific studies have shown that synthetic medical imaging can support dataset augmentation and case generation across ophthalmology, dermatology, radiology, and neuroimaging \cite{yi2019generative,coyner2022synthetic,usman2024brain,rahman2023ambiguous}. Recent foundation-model work has further suggested that these benefits may extend across heterogeneous medical imaging settings \cite{wang2025self}. These developments motivate research-stage generative frameworks that can improve representation before deployment, especially in settings where real subgroup-balanced data remain limited.

However, the promise of AI in medical imaging is constrained by inherent bias in the data \cite{ricci2022addressing, yang2024limits, mihan2024artificial, garin2023medical}. Disparities in healthcare access and disease prevalence produce demographic imbalances across attributes such as skin tone, gender, and age \cite{codella2019skin,shorten2019survey,menze2014multimodal,xu2018fairgan,aggarwal2021diagnostic,tan2019efficientnet,zhang2024fairskin}. These gaps have direct consequences: in the Fitzpatrick 17k dataset, darker skin types (V--VI) account for fewer than 13\% of images \cite{groh2021evaluating}. Disease phenotypes may also manifest differently across populations, rendering certain presentations naturally rarer and more difficult to capture; for example, psoriasis lesions in darker skin often appear grayish or violaceous rather than salmon pink \cite{alexis2014psoriasis}. As a result, models trained on such datasets often underperform for underrepresented groups. Evaluation on the DDI dataset showed 27--36\% drops in diagnostic accuracy for rare skin diseases \cite{daneshjou2022disparities}, while chest radiography models trained with only 5--10\% Black patients showed nearly 50\% lower accuracy in Black subpopulations \cite{norori2021addressing}. Comparable biases have been reported in radiology and neuroimaging \cite{seyyed2020chexclusion,marcinkevics2022debiasing,wachinger2021detect,elyounssi2023uncovering}. Notably, these bias concerns span multiple modalities, yet fairness-ready multimodal medical benchmarks with consistent demographic annotation remain limited. Without targeted interventions, synthetic data risks reinforcing these biases rather than correcting them.

Fairness-aware generative approaches have begun to emerge as a solution. Early work with generative adversarial networks (GANs) introduced fairness objectives but required curated datasets and often degraded image quality \cite{goodfellow2020generative, xu2018fairgan}. For instance, ProstateGAN used a conditional GAN to synthesize realistic prostate diffusion MRI scans stratified by Gleason score, helping mitigate class imbalance, but the augmentation performance still lagged behind that achieved with real data \cite{hu2018prostategan}. More recent efforts have shifted toward diffusion models, which offer greater training stability, higher image fidelity, and improved coverage of long-tailed distributions compared to GANs. Fairness attempts with diffusion have primarily focused on fine-tuning models on balanced datasets \cite{qin2023class,karkkainen2021fairface,li2024fairdiff,teo2023fair}, which is costly, such as FairFace, requiring annotation of more than 100,000 images across gender, age, and race groups \cite{karkkainen2021fairface}, or post-hoc controls on pre-trained models \cite{zhang2023adding,liu2022design,luccioni2023stable,luo2025fairdiffusion}, which may overcorrect or amplify bias: analyses of state-of-the-art diffusion models showed that, even under neutral prompts, 75\%--85\% of outputs were classified as white and 60\%--80\% as male \cite{aldahoul2025ai}. These limitations underscore the need for scalable, modality-agnostic frameworks that exploit the generative strengths of diffusion while embedding fairness objectives directly into the process through attribute-aware resampling, diversity-enhancing regularization, and physician-aligned preference guidance.

Here we present FairGen, a fairness-aware diffusion framework for medical image generation. We study skin images, chest radiography, and T1-weighted MRI because they cover very different medical imaging settings, yet all three have well-documented fairness concerns and still lack a unified multimodal benchmark with consistent demographic annotation. Rather than relying on expensive retraining or fragile post-hoc controls, FairGen builds fairness into the pipeline itself through (i) resampling strategies that rebalance underrepresented attribute--disease pairs, (ii) a diversity-promoting loss that improves subgroup coverage, and (iii) physician-aligned preference guidance that encourages the model to preserve pathology-relevant features in rare and minority cases. Across Fitzpatrick-17k, CheXpert, and OASIS \cite{groh2021evaluating,seyyed2020chexclusion,ioannou2022study}, FairGen lowers Fr\'echet Inception Distance (FID) \cite{heusel2017gans} by 10.1\% for skin images, 13.0\% for chest radiography, and 3.0\% for T1-weighted MRI relative to stable diffusion, while also reducing inter-subgroup FID variance. On downstream benchmark tasks, models trained with FairGen-generated images show sizable fairness gains---95.9\% for skin images, 80.0\% for chest radiography, and 35.2\% for MRI---across demographic parity, equal opportunity, Equalized Sensitive Subgroup Performance, and normalized accuracy range \cite{dwork2012fairness,hardt2016equality,tian2023fairseg}. We further add clinician-facing expert evaluation and external validation on ISIC, IXI, and MIMIC-CXR to test whether these patterns hold beyond the original in-distribution datasets. Taken together, these results position FairGen as a cross-modality framework for fairness-aware medical image synthesis and subgroup-aware model development.


\section*{Results}

\subsection*{Generating Demographically Fair Medical Images Enhances Fairness in AI Diagnostics}

\begin{figure}[H]
    \centering
    \includegraphics[width=\linewidth]{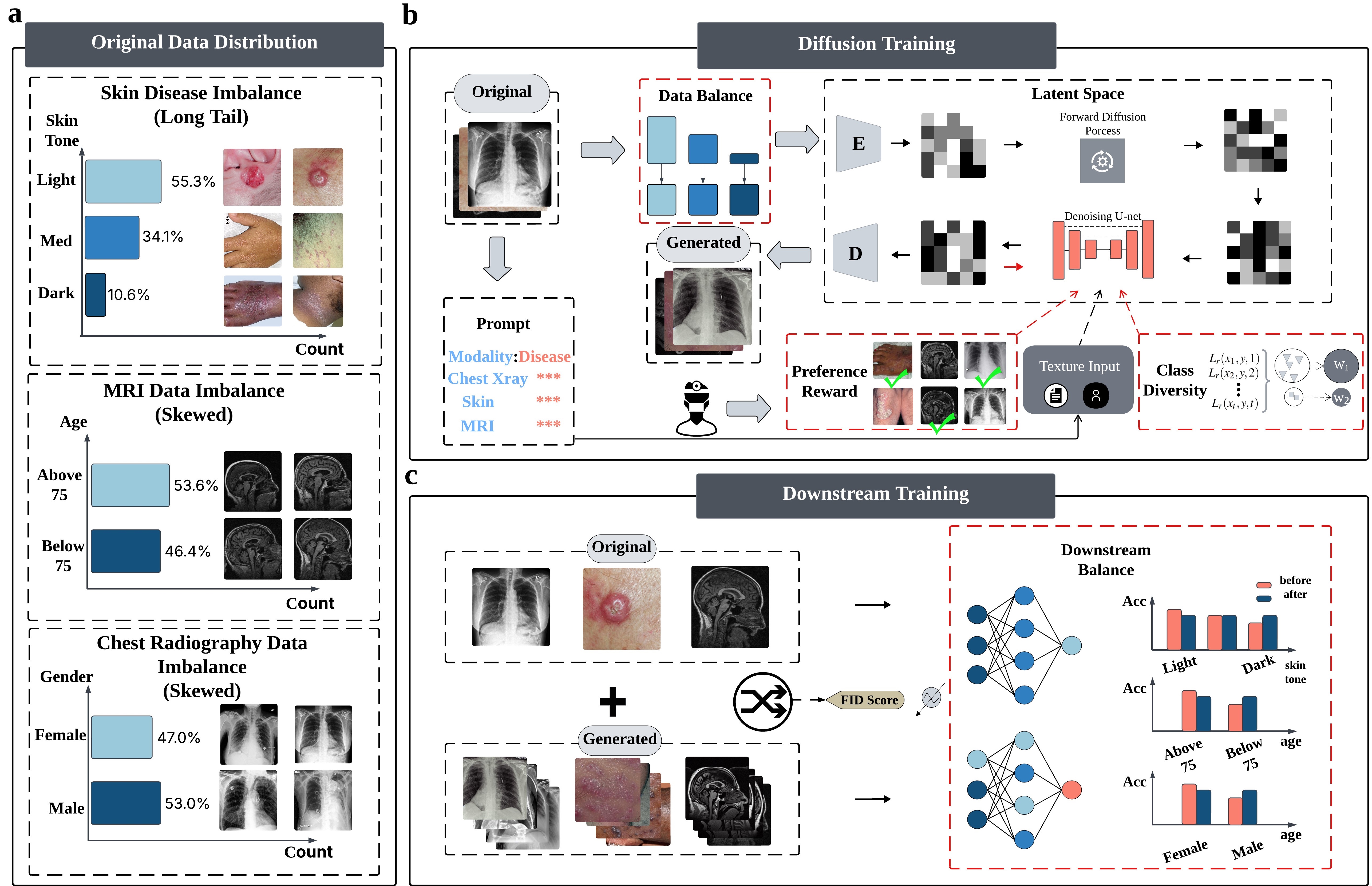}
    \caption{\textbf{Overview of FairGen’s pipeline for mitigating bias through generation and AI-based diagnosis.} 
    \textbf{(a)}, Data imbalance and task structure. Three panels highlight severe demographic disparities: long-tailed skin tone distributions and skewed age/gender ratios in MRI and radiography.
    \textbf{(b)}, FairGen leverages diffusion-based synthesis with data balance, physician-aligned preference rewards, and class diversity to generate balanced images, reducing bias across underrepresented subgroups. (E: encoder transforms images into latent representations; D: decoder reconstructs images from latent representations.) 
    \textbf{(c)}, In the downstream stage, both real and generated data are combined to train classification models, with fairness improvements across sensitive attributes. This integrated workflow effectively addresses long-tail and skewed data problems, promoting fair performance for diverse underrepresented subpopulations.}
    \rule{\textwidth}{0.4pt}
    \label{fig:FairGen_pipeline}
\end{figure}

Our results demonstrate that FairGen, a novel fairness-aware diffusion framework, successfully mitigates demographic biases in medical imaging without compromising downstream classification performance on the benchmark tasks studied here. For readability, we briefly summarize the workflow before turning to the quantitative results: the cohort construction and task definitions are described in Methods under ``Data Setup,'' FairGen training combines resampling, class-diversity regularization, and physician-aligned preference optimization, and the resulting synthetic images are then used for downstream reweighted classifier training. The full pipeline is illustrated in Figure~\ref{fig:FairGen_pipeline}, while the detailed formulations are provided in the Methods section. It generates realistic and diverse images, particularly for underrepresented patient populations and rare diseases, leading to fairer performance in downstream AI-based diagnostic models.

The framework first confronts the widespread data imbalances inherent in clinical datasets together with subgroup-specific phenotype variation (Fig.~\hyperref[fig:FairGen_pipeline]{\ref*{fig:FairGen_pipeline}a}). We highlight three representative cases: skin images exhibit a long-tailed distribution skewed against darker skin tones, where lesions often present with unique phenotypes; T1-weighted MRI scans show underrepresentation of older populations (>${75}$ years), whose brains may show distinct atrophy patterns in dementia; and chest radiographs display a gender imbalance, compounded by the scarcity of rare conditions like pleural effusion.

To counteract these biases at their source, FairGen introduces a fairness-centric generative process (Fig.~\hyperref[fig:FairGen_pipeline]{\ref*{fig:FairGen_pipeline}b}). The pipeline uniquely combines attribute-disease resampling, a latent diffusion model, and two critical fairness-aware signals: (i) a physician-aligned preference reward that guides synthesis towards clinically meaningful and demographically accurate phenotypes, and (ii) a \textbf{class-diversity loss} that ensures distinctiveness and prevents the dominance of majority groups. This process enables FairGen to generate a balanced and faithful set of synthetic images, enriching the representation of rare and minority cases. Ultimately, these high-quality synthetic images are used to augment real data, and a fairness-aware reweighting strategy is applied during classifier training to ensure the generative-stage gains translate into a fair and robust diagnostic model (Fig.~\hyperref[fig:FairGen_pipeline]{\ref*{fig:FairGen_pipeline}c}).

To systematically validate FairGen's efficacy, we conducted a comprehensive evaluation across skin, chest radiography, and T1-weighted MRI modalities. Our assessment was designed to rigorously benchmark FairGen against five state-of-the-art generative baselines---Stable Diffusion (SD)\cite{rombach2022high}, FairDiffusion\cite{luo2025fairdiffusion}, CBRS, SQRS, and CBDM\cite{qin2023class}---at two distinct levels: the generative stage and the downstream diagnostic stage.

At the generative level, we assessed image fidelity using the Fréchet Inception Distance (FID)\cite{heusel2017gans}, where lower scores indicate higher realism. Crucially, we also quantified generative fairness by measuring the variance of FID scores across demographic subgroups, with lower variance signifying fairer image quality. At the downstream diagnostic level, we monitored Classification Accuracy Scores (CAS) to determine whether synthetic augmentation preserved competitive benchmark performance. To confirm the translation of generative improvements into more balanced classifier performance, we employed four established fairness metrics: Demographic Parity (DP)\cite{dwork2012fairness}, Equal Opportunity (EO)\cite{hardt2016equality}, Equalized Sensitive Subgroup Performance (ESSP)\cite{tian2023fairseg}, and Normalized Accuracy Range (NAR)\cite{groh2021evaluating}. Together, these evaluations provide a robust validation of FairGen’s ability to produce high-quality and demographically fair synthetic medical images.

\subsection*{FairGen Outperforms Baselines in Image Fidelity and Realism}

\begin{figure}[H]
    \centering
    \includegraphics[width=\linewidth]{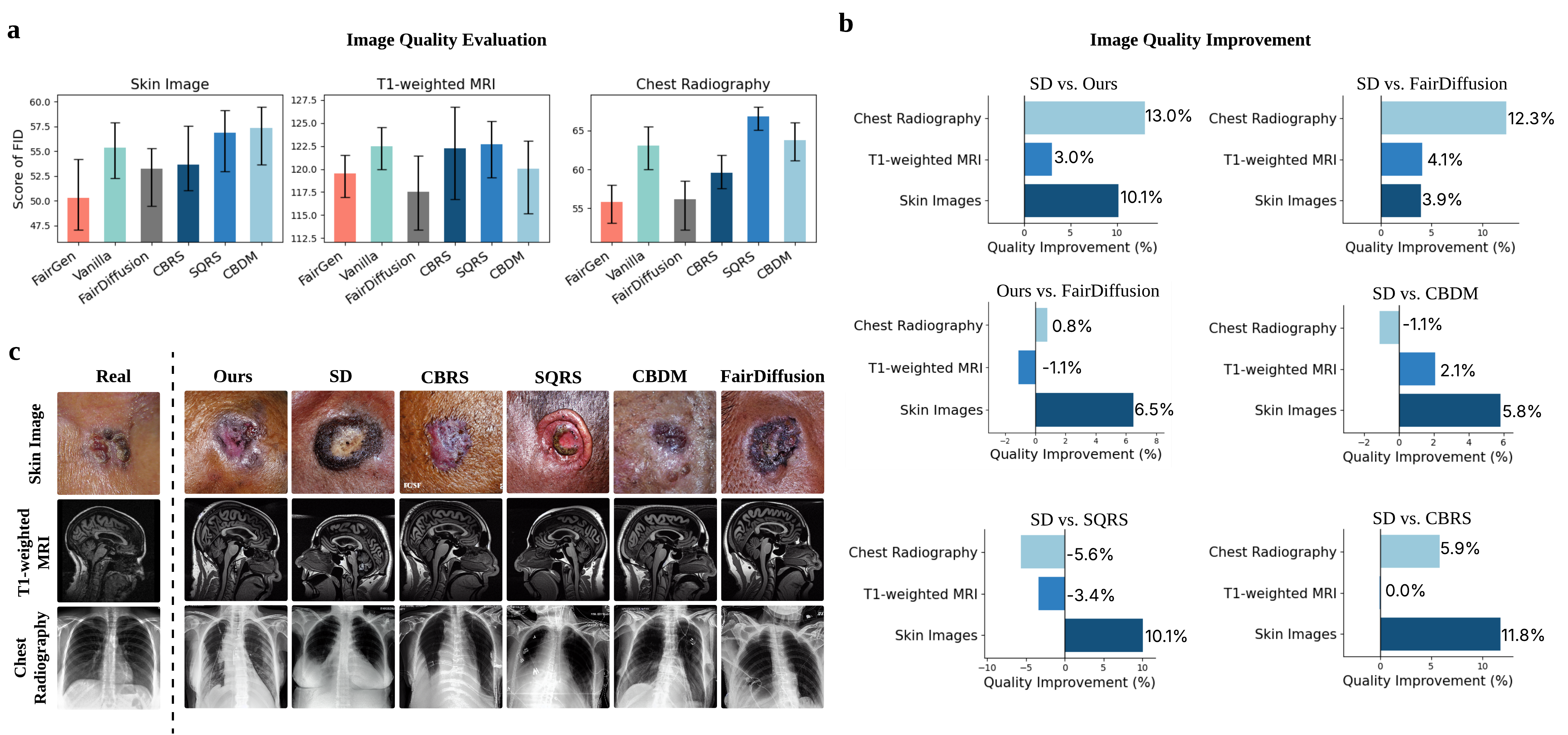}
    \caption{\textbf{Evaluations of synthetic images generated by FairGen compared to baseline methods.} 
    \textbf{(a)}, Quantitative image quality evaluation using Fréchet Inception Distance (FID; lower indicates better quality) comparing FairGen against baseline methods (SD, FairDiffusion, CBRS, SQRS, CBDM) across all modalities.
    \textbf{(b)}, Percentage of relative quality improvement compared to SD baseline, highlighting FairGen’s significant enhancement across all modalities.
    \textbf{(c)}, Visual comparisons demonstrating FairGen’s improved clinical realism and anatomical detail relative to baseline methods and real clinical images.}
    \rule{\textwidth}{0.4pt}
    \label{fig:generation_eval}
\end{figure}

We first assessed the generative quality of FairGen against five state-of-the-art baselines, evaluating its ability to synthesize realistic images across diverse modalities. Quantitatively, FairGen consistently achieved the lowest (best) Fréchet Inception Distance (FID) scores across all three datasets, indicating a higher degree of realism and distributional similarity to the reference data (Fig.~\hyperref[fig:generation_eval]{\ref*{fig:generation_eval}a}). This performance advantage was particularly pronounced for complex or rare presentations. For instance, FairGen outperformed the widely-used SD baseline by a significant margin, with relative FID score improvements of 10.1\% for skin images and 13.0\% for chest radiography (Fig.~\hyperref[fig:generation_eval]{\ref*{fig:generation_eval}b}). These results confirm that embedding fairness objectives does not compromise the model's generative capability; rather, it enhances the fidelity of the synthesized data.

These quantitative gains are mirrored by substantial improvements in anatomical and pathology-relevant detail, as evidenced by visual comparisons (Fig.~\hyperref[fig:generation_eval]{\ref{fig:generation_eval}c}). Specific comparisons reveal the limitations of baseline methods in preserving anatomical fidelity. For instance, in the skin images samples, the SQRS baseline exhibits severe structural hallucinations, generating distorted, cartilage-like features that resemble an ear rather than a cutaneous lesion; conversely, FairGen accurately reconstructs the central ulceration and irregular pigmentation characteristic of the malignancy on darker skin tones. Similarly, in T1-weighted MRI, while methods like CBRS suffer from significant blurring of the gray-white matter interface, FairGen maintains sharp contrast, clearly defining critical landmarks such as the corpus callosum and sulcal patterns. In chest radiography, FairGen synthesizes distinct rib margins and clear lung parenchyma, avoiding the bony structure artifacts observed in CBDM. The clinician-facing experts review reported below further supports that a meaningful subset of FairGen outputs preserve pathology-relevant phenotypes beyond overall visual realism alone.

\subsection*{Experts Evaluation Supports Clinical Feature Preservation Across Modalities}

\begin{figure}[H]
    \centering
    \includegraphics[width=0.95\linewidth]{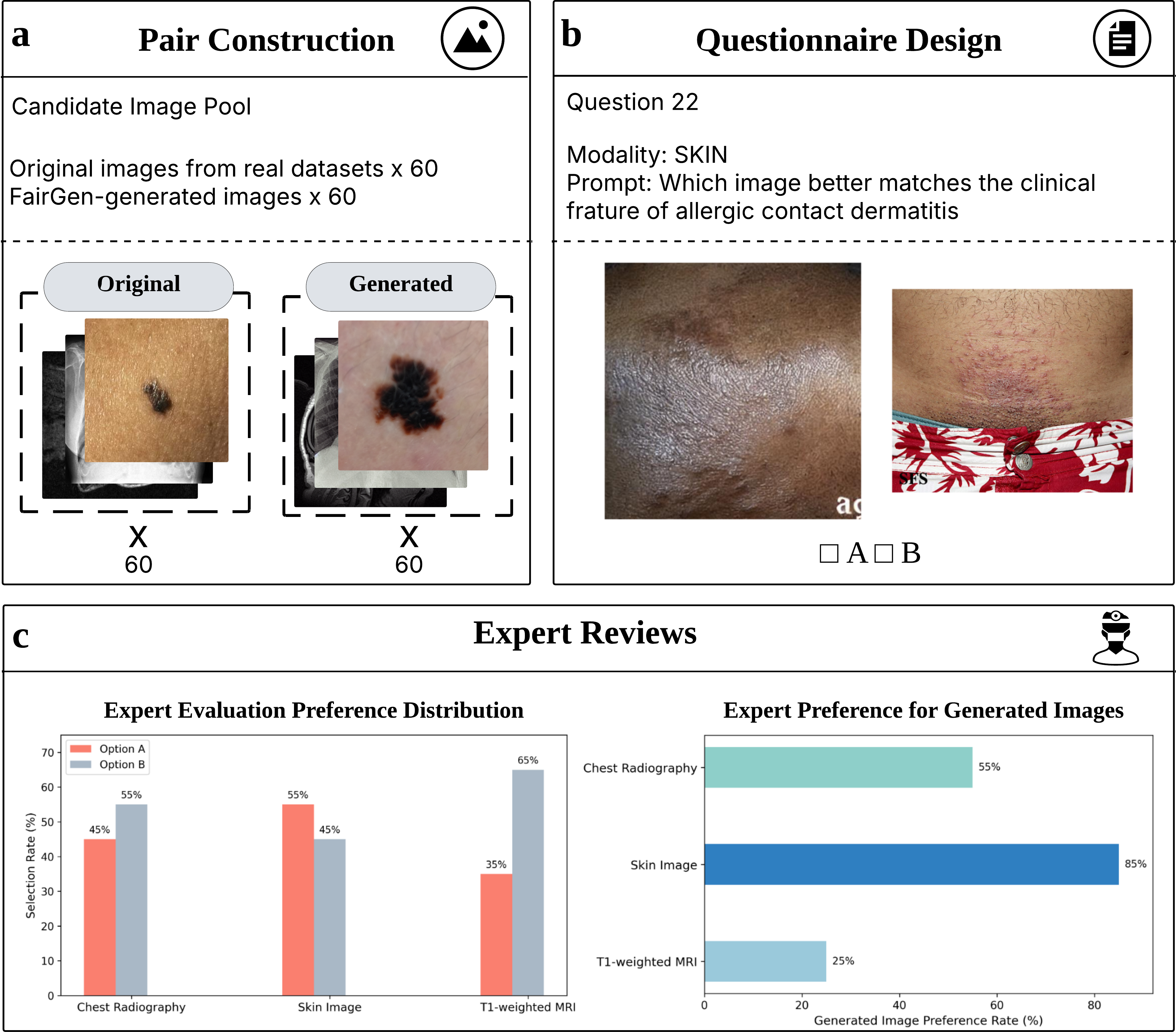}
    \caption{\textbf{Experts evaluation design and outcomes across three medical imaging modalities.}
    \textbf{(a)}, Pair construction for expert review. Candidate images were drawn from original and generated image pools, and modality-specific comparison pairs were constructed for blinded review.
    \textbf{(b)}, Questionnaire design. For each pair, experts were asked to select which image better matched disease-relevant clinical features under a modality-specific prompt.
    \textbf{(c)}, Experts review outcomes. The left panel summarizes the distribution of experts choices across T1-weighted MRI, Skin Image, and Chest Radiography. The right panel summarizes the modality-level preference rates associated with the target comparison setting used in each experts review setup. Together, these results provide supportive clinician-facing evidence that FairGen preserves pathology-relevant visual features beyond standard fidelity metrics.}
    \rule{\textwidth}{0.4pt}
    \label{fig:expert_evaluation}
\end{figure}

To complement FID-based evaluation with experts assessment, we conducted a blinded pairwise evaluation across T1-weighted MRI, Skin Image, and Chest Radiography. As illustrated in Fig.~\ref{fig:expert_evaluation}\textbf{a,b}, candidate images were drawn from original and generated pools and organized into modality-specific comparisons. For each comparison, experts were asked to identify which image better matched disease-relevant clinical features under a standardized modality-specific prompt, rather than simply judging overall visual quality. This design allowed us to assess whether FairGen outputs preserve pathology-relevant information in a way that is recognizable to experts.

The results, summarized in Fig.~\ref{fig:expert_evaluation}\textbf{c}, provide supportive evidence that FairGen preserves clinically meaningful features beyond conventional image-fidelity metrics. Because experts compared generated images against original disease-bearing images, this protocol simultaneously confirmed the pathology-anchor value of the original images and tested whether generated images preserved sufficiently recognizable disease cues to be selected under blinded review. In Chest Radiography, the generated image was preferred in 55\% of comparisons, and in Skin Image, the generated image was preferred in 85\% of comparisons. T1-weighted MRI showed a more conservative experts preference pattern at 25\%, which is consistent with the greater difficulty of assessing subtle dementia-related structural changes from 2D slices alone and the more challenging morphological distinctions in this modality. Together, these findings support the claim that FairGen can preserve pathology-relevant visual cues in experts evaluation, particularly in Chest Radiography and Skin Image.

\begin{figure}[H]
    \centering
    \includegraphics[width=0.95\linewidth]{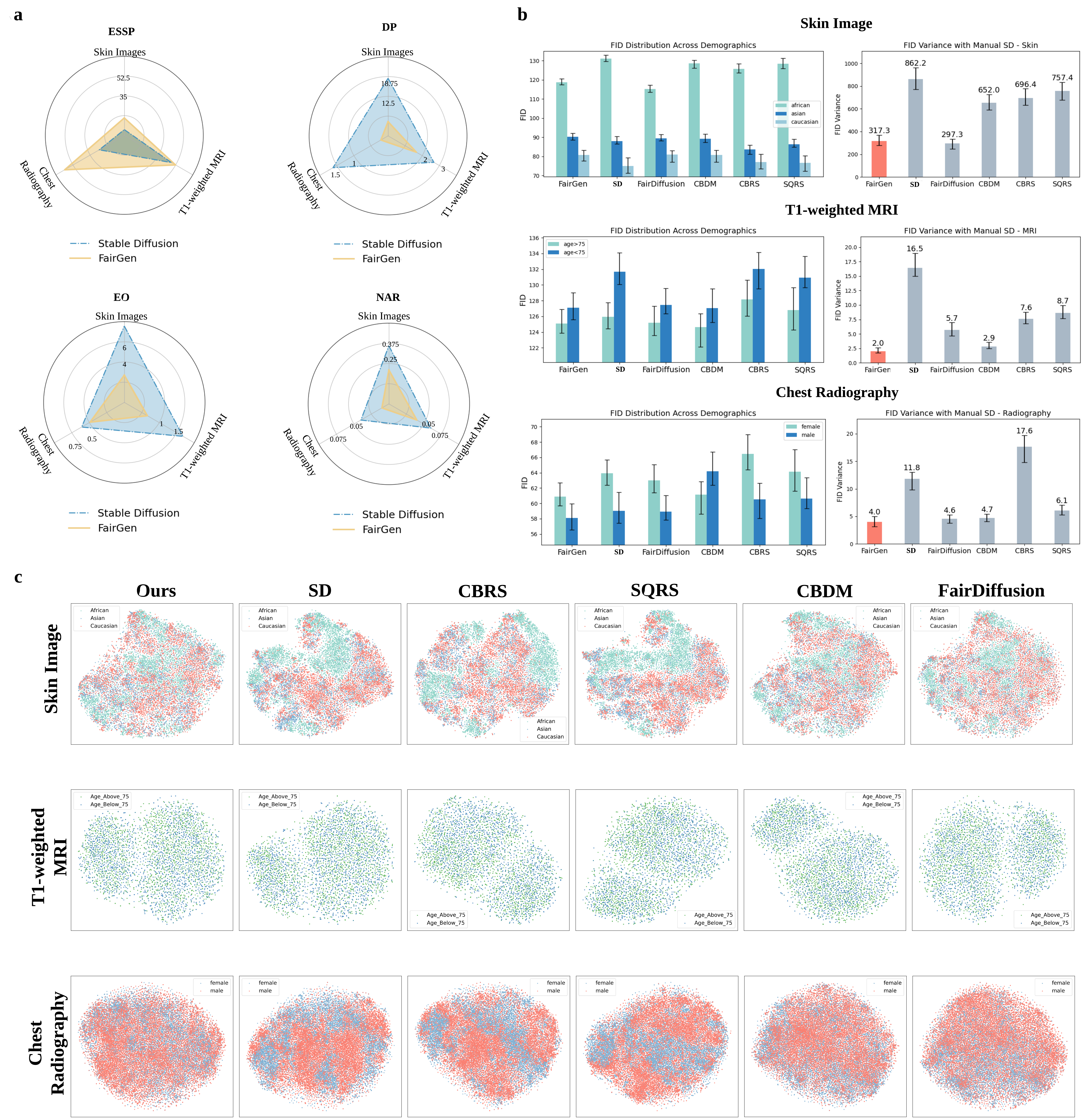}
    \caption{\textbf{Multidimensional assessment of generative fidelity and diagnostic fairness across modalities.} 
    \textbf{(a)}, Downstream diagnostic fairness evaluation using radar charts. FairGen (yellow polygons) consistently outperforms SD (blue polygons) by achieving larger areas for ESSP (higher is better) and more compact, centralized shapes for bias metrics (DP, EO, NAR; lower is better).
    \textbf{(b)}, Quantitative generative fairness assessment via Fréchet Inception Distance (FID). The left columns show FID distribution across demographic subgroups, while the right columns visualize the inter-subgroup FID variance.
    \textbf{(c)}, Qualitative analysis of manifold alignment using t-SNE projections of latent features colored by sensitive attributes. While baseline methods (e.g., SD, CBRS) exhibit distinct clustering or segregation of subgroups (indicating learned bias), FairGen produces a homogenously mixed distribution.}
    \rule{\textwidth}{0.4pt}
    \label{fig:fairness_eval}
\end{figure}

\subsection*{FairGen Mitigates Generative Bias Across Demographic Subgroups}

Beyond improving overall image fidelity, a critical requirement for fair AI is ensuring that high-quality generation extends to all patient populations, not just the majority groups. To evaluate this, we measured the inter-subgroup FID variance, where a lower value indicates fairer image quality across sensitive attributes.

As starkly illustrated in Fig.~\hyperref[fig:fairness_eval]{\ref*{fig:fairness_eval}b}, FairGen dramatically reduced the variance of FID scores among demographic subgroups compared to all baselines, effectively mitigating bias at the generative source. For example, in skin image generation—a modality historically plagued by poor representation of darker skin tones—the FID variance plummeted from 862.2 (SD) to 317.3. Similarly impressive reductions were observed for T1-weighted MRI (from 16.5 to 2.0) and chest radiography (from 11.8 to 4.0). These sharp declines in variance demonstrate that FairGen successfully decouples generation quality from demographic attributes, ensuring that images of minority groups are synthesized with the same high fidelity as those of majority groups. This establishes a robust and unbiased foundation for the downstream diagnostic tasks.

To provide a visual corroboration of these quantitative fairness gains, we utilized t-Distributed Stochastic Neighbor Embedding (t-SNE) to analyze the latent feature distributions of the generated images (Fig.~\hyperref[fig:fairness_eval]{\ref*{fig:fairness_eval}c}). 

The results reveal a striking contrast in how models organize demographic information. As shown in the baseline columns (SD, CBRS, SQRS, etc.), data points are frequently clustered by sensitive attributes---for instance, in the Skin Image modality, samples representing 'African' (green) and 'Caucasian' (orange) skin tones often form distinct, separated islands. This distributional segregation implies that baseline models treat different demographic groups as disjoint manifolds, a clear visual marker of learned bias. 

Conversely, FairGen (Fig.~\hyperref[fig:fairness_eval]{\ref*{fig:fairness_eval}c column 1}) achieves a uniform integration of subgroups. The overlapping color distributions indicate that our framework has successfully aligned the feature manifolds of different populations. By dissolving the artificial boundaries between subgroups, FairGen ensures that the generative process is driven by pathological features rather than demographic biases, providing compelling qualitative evidence for the quantitative reductions in FID variance reported above.

\subsection*{Diagnostic robustness across architectures and optimal backbone selection}

To validate the downstream utility of FairGen, we adopted a two-stage evaluation strategy. First, we benchmarked diagnostic accuracy across diverse machine learning architectures to identify the most performant model. Second, using this optimal architecture as the primary backbone, we assessed whether the fairness benefits of FairGen-generated data could be effectively transferred to the downstream diagnostic task.

We began by analyzing the Classification Accuracy Score (CAS) across four distinct architectures—Vision Transformer (ViT-base-patch16-224), ResNet50, XGBoost, and K-Nearest Neighbors (KNN)—under varying ratios of synthetic data augmentation (Fig.~\hyperref[fig:downstream_eval]{\ref*{fig:downstream_eval}a-d}).

For clarity, the downstream tasks in Fig.~\hyperref[fig:downstream_eval]{\ref*{fig:downstream_eval}a-d} are five-way Skin Image classification (Allergic Contact Dermatitis, Basal Cell Carcinoma, Lichen Planus, Psoriasis, and Squamous Cell Carcinoma), five-way Chest Radiography classification (COVID-19, Edema, Lung Opacity, Pleural Effusion, and Normal), and binary T1-weighted MRI classification (Nondemented vs.\ Demented).

The comparative results reveal two key insights. First, FairGen demonstrates architecture-independent robustness, maintaining competitive accuracy compared to baselines across all four model types. Second, and more importantly, the Vision Transformer (ViT) consistently achieved the relative highest absolute accuracy across all three modalities (Skin, MRI, and X-ray), outperforming both CNN-based (ResNet) and feature-based (XGBoost, KNN) approaches. Based on this finding, we selected ViT as the primary diagnostic backbone for the subsequent fairness evaluation, ensuring that our assessment of fairness is grounded in the strongest-performing model among the tested backbones.

\subsection*{Verifying Fairness Transferability on the High-Performance ViT}

Having established ViT as the superior classifier, we next examined whether FairGen could enhance its fairness without compromising its high accuracy. We computed four established fairness metrics—Demographic Parity (DP), Equal Opportunity (EO), Equalized Sensitive Subgroup Performance (ESSP), and Normalized Accuracy Range (NAR)—specifically for the ViT model.

As presented in Table~\hyperref[tab:fairness_holistic]{\ref*{tab:fairness_holistic}} and visualized in Fig.~\hyperref[fig:fairness_eval]{\ref*{fig:fairness_eval}a}, the results confirm a successful transfer of fairness. In the dermatology task, which exhibits the most severe bias, the FairGen-augmented ViT model achieved an ESSP score of \textbf{15.47} (vs. 5.11 baseline), a three-fold improvement in worst-case subgroup performance. Simultaneously, it reduced the Demographic Parity gap to \textbf{4.51} (vs. 18.22 baseline) and the Equal Opportunity gap to \textbf{2.70} (vs. 7.56 baseline). Similar trends were observed in MRI and X-ray tasks, where NAR was minimized to near-zero levels and significantly optimized the other three fairness metrics. These data demonstrate that FairGen effectively mitigates bias even within the complex, high-capacity decision space of a Vision Transformer.

\definecolor{headerbeige}{RGB}{230, 233, 216}
\definecolor{fairgenhighlight}{RGB}{250, 248, 235}

\begin{table*}[t]
    \centering
    \caption{\textbf{Holistic evaluation of model fairness across different modalities.} Quantitative comparison on Fitzpatrick-17k (skin images), OASIS (T1-weighted MRI), and CheXpert Combine (chest radiography) datasets. We report Equalized Sensitive Subgroup Performance (ESSP, $\uparrow$), Demographic Parity (DP, $\downarrow$), Equal Opportunity (EO, $\downarrow$), and Normalized Accuracy Range (NAR, $\downarrow$). \textbf{Bold} indicates the best performance.}
    \label{tab:fairness_holistic}
    
    \resizebox{\textwidth}{!}{
        \begin{tabular}{l | cccc !{\vrule width 1pt} cccc !{\vrule width 1pt} cccc}
            \toprule
            
            \rowcolor{headerbeige} 
            \textbf{Datasets} & \multicolumn{4}{c !{\vrule width 1pt}}{\textbf{Fitzpatrick-17k}} & 
            \multicolumn{4}{c !{\vrule width 1pt}}{\textbf{OASIS}} & 
            \multicolumn{4}{c}{\textbf{CheXpert Combine}} \\ 
            
            \rowcolor{headerbeige} 
            \textbf{Methods} & ESSP $\uparrow$ & DP $\downarrow$ & EO $\downarrow$ & NAR $\downarrow$ & ESSP $\uparrow$ & DP $\downarrow$ & EO $\downarrow$ & NAR $\downarrow$ & ESSP $\uparrow$ & DP $\downarrow$ & EO $\downarrow$ & NAR $\downarrow$ \\ 
            \midrule
            
            Base (ViT) & 5.11 & 18.22 & 7.56 & 0.36 & 48.24 & 2.68 & 1.68 & 0.06 & 25.62 & 1.61 & 0.61 & 0.04 \\
            Stable Diffusion & 3.76 & 25.04 & 8.19 & 0.41 & 34.18 & 2.87 & 1.87 & 0.06 & 39.21 & 0.86 & 0.43 & 0.02 \\
            CBRS & 5.85 & 15.28 & 6.72 & 0.30 & 31.34 & 3.28 & 2.28 & 0.07 & 35.15 & 1.07 & 0.55 & 0.04 \\
            SQRS & 5.08 & 18.03 & 7.29 & 0.35 & 32.78 & 2.88 & 1.88 & 0.06 & 45.40 & 0.62 & \textbf{0.37} & 0.01 \\
            CBDM & 6.11 & 13.13 & 6.56 & 0.28 & 36.34 & 2.36 & 1.36 & 0.05 & 42.60 & 0.71 & 0.38 & 0.01 \\
            FairDiffusion & 11.16 & 8.13 & 4.52 & 0.25 & \textbf{55.34} & 2.11 & 0.93 & 0.04 & 52.60 & 0.53 & 0.40 & 0.01 \\ 
            
            \rowcolor{fairgenhighlight} 
            \textbf{FairGen (Ours)} & \textbf{15.47} & \textbf{4.51} & \textbf{2.70} & \textbf{0.21} & 52.38 & \textbf{1.66} & \textbf{0.66} & \textbf{0.04} & \textbf{61.08} & \textbf{0.21} & 0.49 & \textbf{0.01} \\ 
            \bottomrule
        \end{tabular}
    }
\end{table*}

\subsection*{Closing the Performance Gap Between Subgroups}

To see how these gains look at the subgroup level, we examined ViT performance in more detail. Fig.~\hyperref[fig:downstream_eval]{\ref*{fig:downstream_eval}e} shows subgroup-specific AUC values as radar charts.

The main pattern is gap reduction rather than simple average-score inflation. Under standard training (SD), the polygons are visibly skewed, with stronger performance on majority groups such as lighter skin or the younger MRI subgroup. With FairGen augmentation, those gaps narrow. In chest radiography, for example, the baseline model favored female patients (0.76 vs. 0.71), whereas FairGen raised the male subgroup from 0.71 to 0.74 and brought the two groups much closer together. In other words, the fairness gain does not come from pushing down the stronger subgroup. It comes mainly from improving the subgroup that was lagging behind.

\subsection*{External Validation Supports Transferability Across Independent Cohorts}

To test whether these trends extend beyond the original datasets, we also evaluated FairGen on three external cohorts drawn from ISIC, IXI, and MIMIC-CXR. These experiments were added in direct response to the concerns about generalizability. Across the three cohorts, FairGen remained competitive in downstream accuracy/AUC and generative fidelity, while the fairness results showed strong ESSP/DP/EO/NAR trade-offs rather than uniform dominance on every metric. We therefore interpret the new external results as evidence of benchmark-level transferability across datasets, not as prospective or deployment-stage clinical validation. Full cohort details and external results are provided in Supplementary Figure 3 and Supplementary Tables 1 and 2.

\subsection*{Embedding Clinical Expertise via Physician-Aligned Preference}

The quantitative results raise a natural question: why does FairGen generate images that are not only visually plausible, but also more useful for subgroup-aware learning? A key part of the answer is the physician-aligned preference dataset used for Direct Preference Optimization (DPO).

To guide the model toward pathology-relevant structure, we incorporated physician input into the annotation process. Fig.~\hyperref[fig:mri_preference_example]{\ref*{fig:mri_preference_example}a} illustrates this protocol. Annotators reviewed pairs of candidate images together with specific clinical prompts, such as "Age above 75, Demented", and selected the image that better matched the target phenotype.

As shown in Fig.~\hyperref[fig:mri_preference_example]{\ref*{fig:mri_preference_example}a}, these choices were based on specific anatomical cues rather than general image quality. In dementia cases, physicians favored images with widened sulcal spaces and cortical atrophy, and rejected samples that had the right metadata but did not show those structural signs. For non-demented elderly cases, they instead looked for narrower sulci and ventricular sizes more consistent with healthy aging. This turns clinical judgment into a structured set of binary preferences, where the chosen image represents the physician-endorsed phenotype.

\subsection*{From Human Annotation to Generative Guidance}

This physician-aligned dataset sits at the center of the FairGen training pipeline. The DPO loss pushes the model away from features that physicians rejected and toward features they preferred. In practice, this means the model is not trained only to match the overall data distribution; it is also nudged toward patterns that clinicians identified as more pathology-relevant. As a result, the generated images are better able to preserve the subtle details highlighted during annotation.

Qualitative review of the final model pointed to two strengths. First, the model was often able to capture specific phenotypes. In neuroimaging examples, physicians noted that FairGen reproduced subtle signs of dementia. For prompts describing older patients, for instance, physicians identified widened sulcal spaces and cortical atrophy in FairGen images, whereas baseline models more often produced generic brain textures without those features.

Second, the physician-aligned dataset appears especially useful for rare conditions. Physicians' feedback suggested that DPO helped preserve structural changes that would otherwise be easy to wash out, such as hippocampal atrophy in demented cases. This supports the idea that the preference mechanism helps prevent phenotype erasure and makes underrepresented conditions easier to preserve during generation.

\begin{figure}[H]
    \centering
    \includegraphics[width=\linewidth]{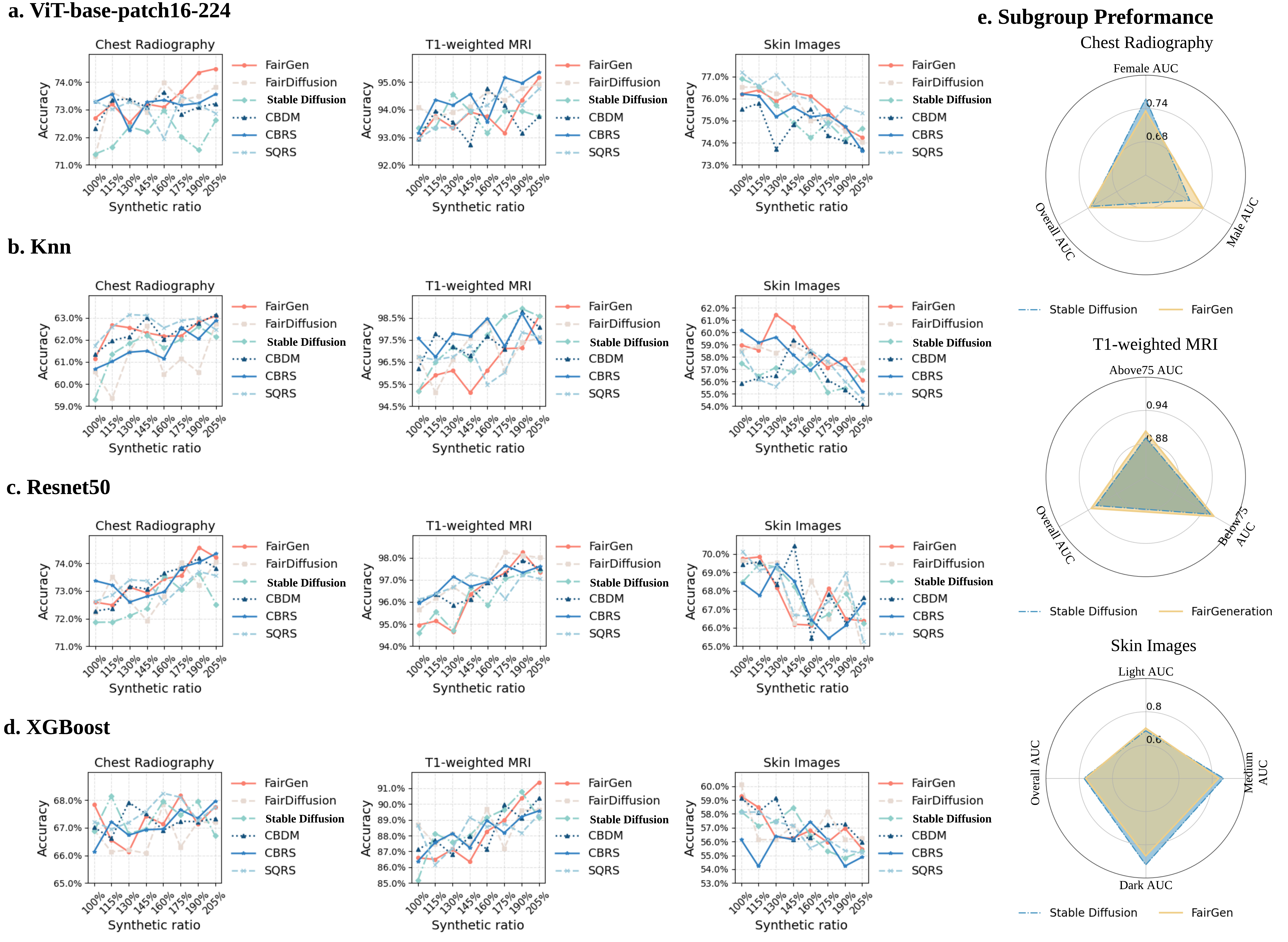}
    \caption{\textbf{Evaluation of diagnostic robustness across diverse classifier architectures and subgroup equity.}
    \textbf{(a-d)}, Diagnostic Classification Accuracy Scores (CAS) under varying synthetic-to-real data ratios across four distinct machine learning architectures: \textbf{(a)} Vision Transformer (ViT-base-patch16-224), \textbf{(b)} K-Nearest Neighbors (KNN), \textbf{(c)} ResNet50, and \textbf{(d)} XGBoost. The downstream tasks are five-way Skin Image classification (Allergic Contact Dermatitis, Basal Cell Carcinoma, Lichen Planus, Psoriasis, Squamous Cell Carcinoma), five-way Chest Radiography classification (COVID-19, Edema, Lung Opacity, Pleural Effusion, Normal), and binary T1-weighted MRI classification (Nondemented vs.\ Demented). 
    FairGen (solid orange line) demonstrates consistent robustness and superior performance compared to five baseline methods, proving that its utility is architecture-agnostic. In stable modalities (MRI, Chest Radiography), FairGen maintains or improves accuracy even at high augmentation ratios.
    \textbf{(e)}, Subgroup-specific performance analysis comparing FairGen against the SD baseline using radar charts. 
    These results are derived from the \textbf{Vision Transformer (ViT) model}, which was selected as the representative architecture due to its superior overall performance compared to other classifiers (KNN, ResNet, XGBoost). The polygons illustrate the Area Under the Curve (AUC) for each sensitive attribute subgroup. FairGen effectively closes the gap for underrepresented populations (e.g., improving 'Dark' skin tone and 'Above 75' age group accuracy) while maintaining high performance for majority groups, resulting in more symmetrical and fairer diagnostic profiles.}
    \rule{\textwidth}{0.4pt}
    \label{fig:downstream_eval}
\end{figure}

\subsection*{Embedding-Space Organization Under Physician-Aligned Guidance}

\begin{figure}[H]
    \centering
    \includegraphics[width=0.95\linewidth]{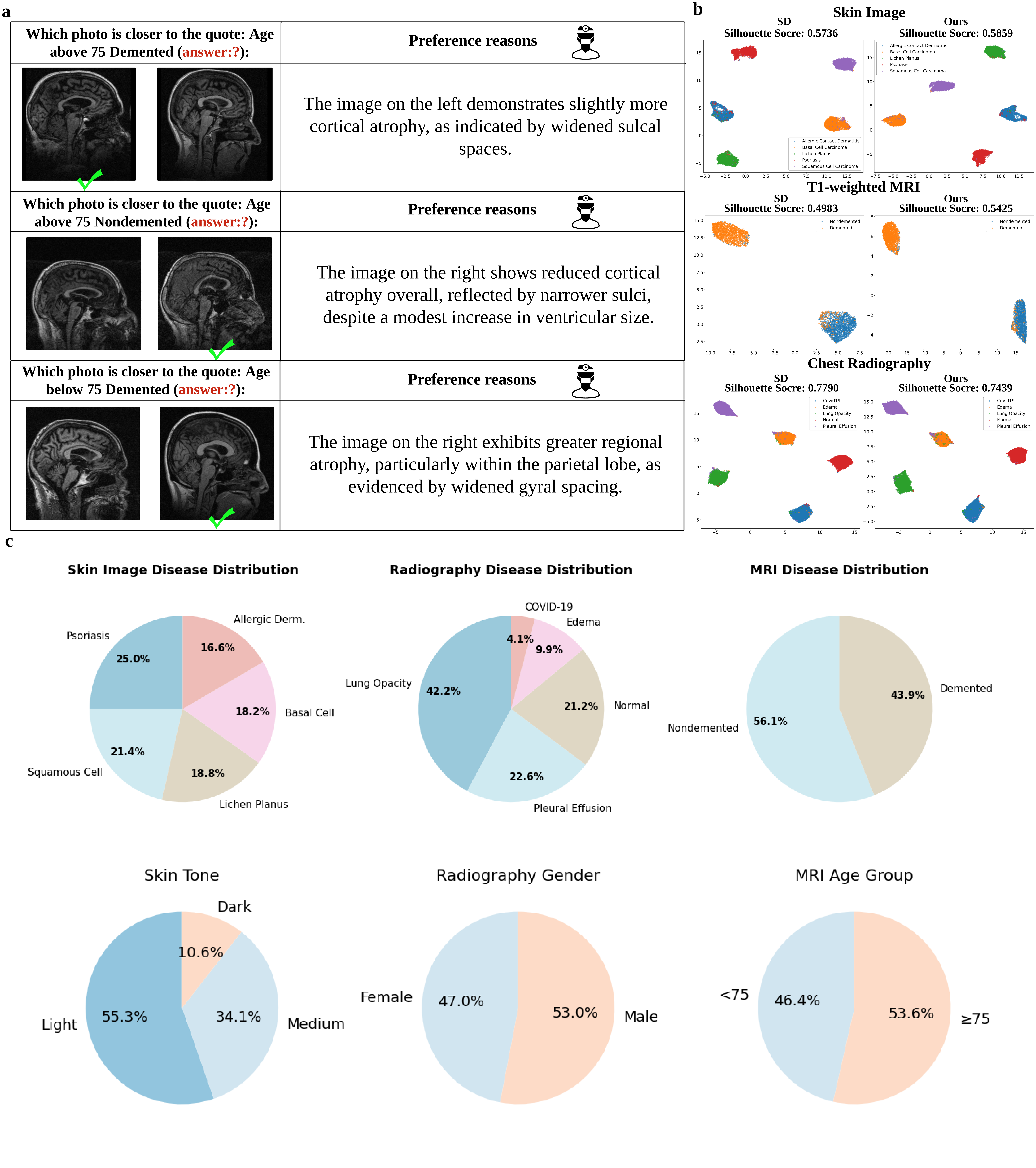}
    \caption{\textbf{Evaluation of expert alignment, latent representations, and data characteristics.} 
    \textbf{(a)}, Physician preference annotation. Experts selected images based on anatomical fidelity (e.g., cortical atrophy) to serve as ground truth for Direct Preference Optimization (DPO). 
    \textbf{(b)}, UMAP projections of latent features, colored by disease labels, with silhouette scores reported above each panel. The visualization is used as a complementary view of embedding-space organization rather than a standalone performance metric. 
    \textbf{(c)}, Original dataset distributions. Pie charts quantify the severe imbalance across sensitive attributes (e.g., skin tone, age) and disease classes in the training data.}
    \label{fig:mri_preference_example}
    \rule{\textwidth}{0.4pt}
\end{figure}

To better understand the downstream fairness gains, we analyzed the learned embedding space. We visualized two-dimensional UMAP projections of the test-set embeddings, colored by disease label, and added silhouette scores as a compact summary of within-class compactness and between-class separation (Fig.~\hyperref[fig:mri_preference_example]{\ref*{fig:mri_preference_example}b}). In simple terms, a higher silhouette score means samples tend to lie closer to their own class than to the nearest alternative class.

The plots show that the effect is modality-dependent. In the baseline model (SD), disease clusters are visible, but in some modalities their boundaries are diffuse and overlap substantially. This is especially noticeable for MRI, where the unguided model appears to produce a noisier feature space.

For Skin Image and T1-weighted MRI, FairGen increases the silhouette score from 0.5736 to 0.5859 and from 0.4983 to 0.5425, respectively, suggesting tighter disease-level organization after physician-aligned guidance. In Chest Radiography, the score falls from 0.7790 to 0.7439, so we do not read this figure as a uniform improvement in separability across all modalities. Instead, we view the UMAP-plus-silhouette analysis as a complementary window into the mechanism: in some settings FairGen sharpens disease clusters, while in others it may trade some cluster compactness for a broader redistribution of subgroup information. We therefore use Fig.~\hyperref[fig:mri_preference_example]{\ref*{fig:mri_preference_example}b} as an interpretability aid rather than as a standalone performance claim.

\section*{Discussion}



FairGen was designed to address demographic imbalance and rare disease scarcity across three very different imaging settings: skin images, chest radiography, and T1-weighted MRI. Our results show that physician-aligned guidance can help the model preserve subgroup-specific and rare phenotypes while improving fairness on downstream benchmark tasks. In practical terms, the generated data are most useful as a supplement to existing datasets, especially when real samples are scarce or unevenly distributed because of disease rarity or broader healthcare disparities~\cite{obermeyer2019dissecting}. We adds clinician-facing expert review and external validation on ISIC, IXI, and MIMIC-CXR, which together show that the main findings are not limited to standard fidelity metrics or to the original in-distribution cohorts. Taken together, these results support FairGen as a research-stage framework for fairness-aware image generation and subgroup-aware model development.

A consistent result across the three modalities is that fairness improved without a large drop in diagnostic accuracy. In several settings, classifiers trained with FairGen-augmented data matched or exceeded the accuracy of the baseline models. This matters because fairness interventions are often assumed to come with a sizable performance penalty~\cite{wachinger2021detect}, particularly in settings linked to underdiagnosis for underserved populations~\cite{seyyed2021underdiagnosis}. Earlier fairness-aware generative approaches have typically relied either on balanced fine-tuning, which is costly and annotation-heavy, or on post-hoc controls, which can overcorrect or amplify bias~\cite{karkkainen2021fairface,aldahoul2025ai}. More broadly, medical AI systems are also known to latch onto shortcut correlations that do not hold up across subgroups or institutions~\cite{geirhos2020shortcut,degrave2021ai}. FairGen takes a different route by combining resampling, class-diversity regularization, physician-aligned guidance, and downstream reweighting. In our experiments, that combination improved EO~\cite{hardt2016equality}, DP~\cite{dwork2012fairness}, ESSP~\cite{tian2023fairseg}, and NAR~\cite{groh2021evaluating} while keeping predictive performance competitive. The overall picture is that subgroup equity can be improved without paying a large accuracy cost.

Physician feedback was most useful in clarifying what FairGen was actually preserving. In the neuroimaging setting, the collaborating physician emphasized that structural MRI remains an important modality for studying late-life neurodegeneration, even though definitive clinical workups often rely on more specialized biomarkers such as PET-based measures of amyloid-$\beta$ and phosphorylated tau-217~\cite{jack2018nia,chen2022pan}. Within that context, the physician-guided preferences appeared to help the model preserve subtle anatomical cues such as widened sulcal spaces and cortical atrophy, rather than defaulting to generic brain textures. We view this as evidence that FairGen is clinically grounded at the level of dataset augmentation, educational illustration, and subgroup-aware model development. It should not be read as evidence that the method is ready to replace existing workflows or serve as a deployment-stage decision-support system.

Several limitations should be acknowledged. First, the added expert study was designed as a blinded clinician-facing pairwise review rather than a formal multi-rater agreement study, so inter-rater reliability statistics such as ICC are not reported. Second, although external validation on ISIC, IXI, and MIMIC-CXR substantially strengthens the manuscript, these analyses remain benchmark-level evaluations rather than prospective clinical validation. Third, the external MRI experiment on IXI evaluates sex-aware age-group prediction rather than a direct replication of the original dementia task, reflecting the limited availability of public MRI datasets that jointly provide disease labels and sensitive-attribute annotations. Therefore, more exhaustive calibration, error-severity, and clinical-risk analyses remain important directions for future work. Finally, the risk of using AI is another limitation of the current study. Synthetic medical images can look plausible while carrying unsupported pathology, altered anatomy, misleading subgroup cues, or shortcut correlations that fail under domain shift~\cite{ali2025enhancing,obuchowicz2026will,kompa2021second,degrave2021ai,bergquist2025artificial}. Because demographic bias can remain hidden even when average accuracy is strong~\cite{banerjee2021reading,seyyed2021underdiagnosis,ricci2022addressing,yang2024limits,xu2024addressing}, FairGen should be interpreted as a research-stage augmentation and fairness-evaluation tool that still requires prospective validation, subgroup-specific calibration, human oversight, and post-deployment monitoring before clinical use.


Several next steps follow naturally from these results. One is to test FairGen in additional medical domains, such as computational pathology, where both data scarcity and demographic imbalance remain serious practical problems~\cite{lu2021data}. Another is to move beyond the current 2D setting and evaluate the framework on richer imaging contexts, including 3D MRI volumes and multimodal diagnostic tasks~\cite{acosta2022multimodal}. 


The evaluation itself also needs to be strengthened. Larger clinician studies, formal multi-rater agreement analysis, broader external validation, and fairness analyses that include calibration, error severity, and clinical risk would all make the evidence base stronger. It would also be useful to test whether the same ideas carry over to newer medical foundation models and to privacy-preserving settings such as federated learning~\cite{moor2023foundation,kaissis2020secure}. We view FairGen as one step toward more equitable pre-deployment medical AI research, not as a finished clinical solution.

\section*{Methods}

\subsection*{Data Setup}

Biases in medical imaging datasets pose significant challenges to achieving equitable classification performance, particularly across populations defined by sensitive demographic attributes. In this study, we examine three representative imaging modalities—skin images, chest radiography, and T1-weighted MRI scans—and define modality-specific sensitive attributes based on established clinical standards. These three modalities were selected because they span heterogeneous medical-imaging settings in which fairness concerns have been repeatedly documented, while a single widely adopted multimodal benchmark with consistent demographic annotation remains limited.

For skin images, sensitive attributes are categorized using the Fitzpatrick skin tone scale (light, medium, dark), which is widely adopted in dermatological studies. We utilize the open-source Fitzpatrick17k dataset\cite{groh2021evaluating}, selecting 2,150 images spanning five disease types, including both common conditions (allergic contact dermatitis, basal cell carcinoma, psoriasis, and squamous cell carcinoma) and a rarer inflammatory disorder, lichen planus. 

For chest radiography, the sensitive attribute in the main in-distribution study is binary gender (female vs.\ male), reflecting strong sex-related anatomical variation in thoracic radiography and allowing a consistent subgroup definition within the original training data. We curate a combined dataset comprising 13,646 images from the CheXpert dataset\cite{irvin2019chexpert} and 576 images from the COVID-19 dataset\cite{cohen2020covid}, covering frequent findings such as lung opacity and edema, as well as rare cases such as pleural effusion. A separate race-based external chest-radiography analysis is additionally reported in the Supplementary Figure 3 and Supplementary Table 2.

For T1-weighted brain MRI, we use age-based bands (\(<75\) vs.\ \(\geq75\)) to account for structural changes in aging brains. This threshold was chosen as a clinically interpretable late-life division and also maintains analyzable subgroup sizes in the OASIS cohort. Images are drawn from the OASIS dataset\cite{marcus2010open}, yielding 1,234 axial-slice images from 150 subjects, including both non-demented and demented cases, the latter representing a rarer subgroup within neuroimaging cohorts.

Each modality is associated with a classification task:
\begin{itemize}
    \item \textbf{Skin Images:} Allergic Contact Dermatitis, Basal Cell Carcinoma, Lichen Planus, Psoriasis, and Squamous Cell Carcinoma.
    \item \textbf{Chest Radiography:} COVID-19, Edema, Lung Opacity, Pleural Effusion, and Normal.
    \item \textbf{T1-weighted MRI Scans:} Nondemented vs.\ Demented states.
\end{itemize}

These label definitions are used consistently throughout the main-text downstream evaluations, including Fig.~\ref{fig:downstream_eval} and Table~\ref{tab:fairness_holistic}. External validation tasks on ISIC, IXI, and MIMIC-CXR are defined separately in the Supplementary Table 1 because they are constrained by the available labels and sensitive-attribute annotations in each external cohort.

We denote the dataset as \({S} = \{(x_i, r_i, d_i)\}_{i=1}^N\), where \(x_i \in {X}\) is the input image, \(r_i \in {R}\) is the sensitive attribute (e.g., skin tone, gender, or age group), and \(d_i \in {D}\) is the disease label.

Fig.~\hyperref[fig:mri_preference_example]{\ref*{fig:mri_preference_example}c} summarizes the distribution of image samples across disease classes for each modality. To provide a clearer view of demographic imbalance in different sensitive attributes, Fig.~\hyperref[fig:mri_preference_example]{\ref*{fig:mri_preference_example}c} visualizes the overall composition of each attribute subgroup as pie charts. The results indicate severe underrepresentation of dark skin tones in dermatology datasets (only 10.6\%), and moderate but non-negligible imbalance in gender (X-ray) and age groups (MRI). These disparities underscore the necessity of fairness-aware modeling and data generation strategies for downstream diagnostic tasks.

\subsection*{Ethics statement}
All analyses in this study were conducted in accordance with the Declaration of Helsinki. The present work used only retrospective, de-identified, publicly available imaging datasets. No new participants were recruited and no new clinical data were collected for this study.

The ethics and consent status of the source datasets follows the corresponding public dataset documentation and source publications. In the main in-distribution experiments, Fitzpatrick17k is a benchmark assembled from publicly available dermatology image sources rather than a single newly recruited cohort, so the public release does not provide one unified study-specific institutional review board approval number for the benchmark itself. The COVID-19 Image Data Collection is likewise a public compilation of images assembled from publicly available publications and web sources rather than a single prospectively enrolled clinical cohort. For CheXpert and OASIS, we used only the released, de-identified public data under the corresponding dataset terms; the original cohort-level approval and consent procedures are described in the respective source publications and access documentation. In the external validation analyses, ISIC and IXI were likewise used under their public release terms; where the public dataset documentation did not report a single dataset-level approval number in a consistent way, we do not infer one here. For MIMIC-CXR, the public source documentation states that the database was approved by the Institutional Review Boards of Beth Israel Deaconess Medical Center and the Massachusetts Institute of Technology, and that individual patient consent was waived because the data were de-identified and did not affect clinical care.

Accordingly, the current study should be understood as a secondary analysis of previously released, de-identified public data. Under these conditions, no additional local institutional review board approval or new informed-consent procedure was required for the present analysis.

\subsection*{Training Data Resampling}
Resampling plays a critical role in mitigating data imbalance before feeding the dataset into the training pipeline. In the \textit{FairGen} framework, we employ two complementary resampling strategies: \textbf{Class Balanced Random Sampling (CBRS)} and \textbf{Square Root Random Sampling (SQRS)}. These methods aim to ensure that the training data maintains a balanced representation across both sensitive attributes and disease classes.



\paragraph{Class Balanced Random Sampling (CBRS).}
CBRS is designed to counteract data imbalance by assigning larger weights to underrepresented attribute–disease groups. To improve its adaptivity, we introduce a gradient-aware mechanism to refine weight assignment based not only on data frequency but also on the optimization status of each group. The enhanced sample weight for a group $(r,d)$ is defined as:
\begin{equation}
w_{r,d} = \frac{1}{N_{r,d}} \cdot \left(1 + \alpha \cdot \frac{\overline{\|\nabla {L}_{r,d}\|}}{\sum\limits_{(r',d')}\overline{\|\nabla {L}_{r',d'}\|}} \right)
\label{eq:CBRS_grad}
\end{equation}
where \(N_{r,d}\) is the number of training samples in group $(r,d)$, \(\overline{\|\nabla {L}_{r,d}\|}\) denotes the average gradient norm of the loss function for this group (computed over a recent window of mini-batches), and \(\alpha\) controls the influence of gradient scaling. This formulation retains frequency-based balancing while adaptively increasing the weights of groups that yield higher training gradients, reflecting learning difficulty or under-optimization. In effect, it encourages the model to pay more attention to both minority and harder-to-learn groups during training.

\paragraph{Square Root Random Sampling (SQRS).}
SQRS adopts a more conservative reweighting scheme by applying a square-root transformation to group frequencies, avoiding over-amplification of rare categories. We extend SQRS similarly with a gradient-aware adjustment, resulting in the following weight formulation:
\begin{equation}
w_{r,d} = \frac{1}{\sqrt{N_{r,d}}} \cdot \left(1 + \alpha \cdot \frac{\overline{\|\nabla {L}_{r,d}\|}}{\sum\limits_{(r',d')}\overline{\|\nabla {L}_{r',d'}\|}} \right)
\label{eq:SQRS_grad}
\end{equation}
This approach smooths the impact of class imbalance while integrating dynamic feedback from gradient norms to emphasize underperforming or difficult groups. Compared to CBRS, the square-root base ensures less aggressive weighting, which is particularly beneficial in scenarios where extreme over-sampling may introduce noise.

\subsection*{Backbone Architecture, Complexity, and Reproducibility}
At the network level, FairGen reuses the latent diffusion backbone of Stable Diffusion v1-4 rather than introducing a separate image generator. The implementation loads four standard components from the pretrained checkpoint: a DDPM scheduler, a CLIP tokenizer/text encoder for prompt conditioning, an AutoencoderKL (VAE) for mapping images to and from latent space, and a conditional U-Net denoiser. During fine-tuning, the VAE and text encoder remain frozen, while the U-Net is the primary trainable image-generation module. In this sense, FairGen and the Stable Diffusion reference model share the same base text-to-image architecture.

What changes in FairGen is the training strategy. The resampling stage modifies how attribute--disease pairs are presented to the model. The balanced diffusion stage adds fairness-aware regularization on top of the standard denoising objective. In the code-faithful implementation, the class-diversity branch evaluates the same noisy latent under an alternative subgroup-conditioned prompt, producing an additional regularization/comparison term on top of the original DDPM loss. When physician-aligned preference guidance is enabled, an extra preference-learning signal is added during training through prompt-conditioned preferred/non-preferred image pairs. These additions alter the optimization objective, but they do not replace the Stable Diffusion backbone itself.

This distinction is important for computational complexity. For the Stable Diffusion v1-4 checkpoint used in our experiments, the text encoder contains 123,060,480 parameters, the VAE contains 83,653,863 parameters, and the U-Net contains 859,520,964 parameters, for a total of 1,066,235,307 parameters. FairGen reuses this same backbone parameterization. Therefore, inference-time complexity remains effectively the same order as Stable Diffusion, because the fairness-specific losses are used during training rather than inserted into the sampling path. The main computational overhead appears during training: the balanced comparison branch adds one extra text-encoder pass and one extra U-Net forward pass per iteration, so the dominant cost is approximately one additional U-Net evaluation relative to vanilla Stable Diffusion. When physician-aligned preference guidance is active, there is further training-time overhead from the preference-learning branch, but again without increasing the size of the image-generation backbone itself.

To make the practical training burden explicit, all major diffusion experiments were implemented with Hugging Face Diffusers and Accelerate. Training was performed on NVIDIA Tesla V100 32GB GPUs using fp16 mixed precision and gradient checkpointing. For the main diffusion fine-tuning runs, we used a per-device batch size of 8 with 4-step gradient accumulation, yielding an effective batch size of 32, together with a learning rate of \(3\times10^{-7}\), a cosine scheduler, and 15{,}000 optimization steps. In our current experiments, representative wall-clock training times ranged from approximately 18.7 to 32.7 hours per diffusion model depending on the method variant. This setup reflects the fact that FairGen shares the same Stable Diffusion v1-4 backbone as the reference model and incurs additional cost primarily through fairness-specific auxiliary training branches rather than through a larger inference-time network.

\subsection*{Balanced Diffusion Model Training}  
To address the intrinsic class imbalance in medical imaging datasets, we introduce a balanced training strategy within the diffusion modeling framework. This approach establishes a foundational mechanism to incorporate fairness constraints into the latent representation space, enabling the generation of high-fidelity synthetic medical images with equitable representation across diverse demographic attributes. This approach consists of two core components: the foundational training objective of the diffusion model and the inclusion of a class diversity loss term to mitigate biases during the generative process.

\paragraph{Original Diffusion Model Training.}  
The diffusion model operates by reversing a predefined noise addition process to reconstruct high-fidelity images from random noise\cite{rombach2022high}. The training objective is designed to minimize the discrepancy between reconstructed images and real data samples, thereby learning a generative model capable of accurately representing the dataset's underlying distribution.

The loss function for the diffusion model is based on the Variational Lower Bound (VLB) and can be expressed as:  
\begin{equation}
L_{\text{DM}} = \mathbb{E}_{x, t, \epsilon} \big[ \|\epsilon - \epsilon_\theta(x_t, t)\|^2 \big]
\label{eq:diffusion_loss}
\end{equation}
where \(x\) denotes the original data sample, \(\epsilon\) is Gaussian noise added to \(x\), and \(x_t\) represents the noised version of \(x\) at timestep \(t\). The term \(\epsilon_\theta(x_t, t)\) is the predicted noise at timestep \(t\), parameterized by the neural network \(\theta\). The model parameters \(\theta\) are optimized to predict the noise \(\epsilon\) accurately, allowing the reverse diffusion process to reconstruct the original image.

While this training objective is effective for generating high-quality images, it does not explicitly address the issue of class imbalance. Consequently, the resulting synthetic data may underrepresent minority classes, potentially introducing biases into downstream tasks.

\paragraph{Promoting Class Diversity Loss.}  
To mitigate the biases arising from class imbalances, a class diversity loss term is integrated into the diffusion model training. This extension ensures the generation of diverse synthetic samples across all sensitive attributes and disease classes, promoting equitable representation in the output distribution.

We introduce Class-Balancing Square Root (CBSQ) incorporates a class diversity loss term, \({L}_r\), which is defined as: 
\begin{equation}
{L}_r(x_t, y, t) = \frac{t}{|{Y}|} \sum_{y' \neq y} \|\epsilon_\theta(x_t, y) - \epsilon_\theta(x_t, y')\|^2
\label{eq:diffusion_balance_loss}
\end{equation}
where \(\epsilon_\theta(x_t, y)\) denotes the estimated noise for the noisy image \(x_t\) conditioned on class label \(y\), and \(y'\) represents a randomly sampled class label distinct from \(y\). The term \(t\) acts as a scaling factor proportional to the diffusion timestep, while \(|Y|\) is the total number of classes in the dataset.

This loss (Eq.~\ref{eq:diffusion_balance_loss}) penalizes the model when noise predictions across different classes are inconsistent or indistinguishable, thereby maintaining clear distinctions between classes during training. By enforcing this constraint, the model is encouraged to generate samples that are representative of the entire class distribution, mitigating mode collapse and overrepresentation of majority classes.

\paragraph{Preference Reward.}  
To align the diffusion model outputs with Physician-aligned preferences while ensuring equitable representation across sensitive attributes and disease classes, we incorporate a preference-based reward mechanism inspired by Direct Preference Optimization (DPO)~\cite{rafailov2023direct}. Specifically, we first construct a reward model \( r_{\phi}(a, d, x_0) \) by leveraging a human-annotated dataset of preference pairs defined as:
\begin{equation}
{D} = \{(a, d, x^w_0, x^l_0)\}
\label{eq:reward_dataset}
\end{equation}
where each tuple comprises attribute \(a\), disease class \(d\), and two generated samples—\( x^w_0 \) (preferred) and \( x^l_0 \) (non-preferred). In practice, each reviewed item is stored as a prompt-conditioned preference tuple: a clinically meaningful text prompt, a preferred image, and a non-preferred image. Physician annotators select which candidate image better matches the pathology-relevant features described by the prompt, rather than simply choosing the image that appears most realistic.

The prompt templates are modality-specific and disease-focused. For example, the annotation prompt fixes the relevant attribute--disease condition and asks which image more faithfully matches the target presentation under that condition. This design allows the preference signal to encode clinically grounded winner/loser comparisons that can be passed directly into the DPO objective. We emphasize that these training-stage physician preference pairs are distinct from the separate clinician-facing expert evaluation, which is reported later as a validation analysis rather than used as DPO supervision.

The reward model \( r_{\phi}(a, d, x_0) \) is optimized to accurately distinguish preferred from non-preferred samples through the binary preference objective:
\begin{equation}
L_{\text{RM}}(\phi) = -\mathbb{E}_{(a,d,x^w_0,x^l_0)\sim{D}}\left[\log\sigma\bigl(r_{\phi}(a, d, x^w_0) - r_{\phi}(a, d, x^l_0)\bigr)\right]
\label{eq:reward_loss}
\end{equation}
where \(\sigma(\cdot)\) denotes the sigmoid function, ensuring the reward model assigns higher scores to preferred images consistently.

To guide the diffusion model towards generating fair and clinically preferred images at each intermediate timestep \(t\), we define a timestep-specific reward function:
\begin{equation}
r_{\theta}(a, d, x_t) = r_{\phi}(a, d, x_0) - \beta\|\epsilon_{\theta}(x_t, t)-\epsilon\|^2_2
\label{eq:reward_function}
\end{equation}
where \(\epsilon_{\theta}(x_t, t)\) represents the diffusion model's predicted noise at timestep \(t\), \(\epsilon\) is the actual sampled noise drawn from the standard Gaussian distribution, and \(\beta\) serves as a balancing hyperparameter that penalizes deviations from the standard diffusion objective.

Consequently, the overall DPO loss function for the diffusion training integrates these per-timestep rewards and is formalized as:
\begin{equation}
L_{\text{DPO}}(\theta) = -\mathbb{E}_{(a,d,x^w_t,x^l_t)\sim{D},\,t}\left[\log\sigma\Bigl(\lambda\cdot\bigl(r_{\theta}(a, d, x^w_t)-r_{\theta}(a, d, x^l_t)\bigr)\Bigr)\right]
\label{eq:reward_loss_time_step}
\end{equation}
where \(\lambda\) scales the relative influence of the reward difference between preferred and non-preferred synthetic samples (see Supplementary Figures 1 and 2 for details). 
In summary, the preference-based dataset (Eq.~\ref{eq:reward_dataset}) establishes the foundation for learning physician-aligned preferences, enabling the optimization of the reward model (Eq.~\ref{eq:reward_loss}) to reliably distinguish preferred samples. This optimized reward model subsequently guides the definition of the timestep-specific reward function (Eq.~\ref{eq:reward_function}), ensuring diffusion model outputs align closely with clinical relevance and fairness considerations at each intermediate step. Collectively, these components underpin and directly inform the formulation of the final Direct Preference Optimization loss (Eq.~\ref{eq:reward_loss_time_step}), effectively embedding human expert judgments and fairness constraints into the diffusion training process, thus significantly enhancing both the clinical realism and equitable representation of synthetic medical images.

\subsection*{Downstream Balancing}
After training the diffusion model, we integrate the generated debiased synthetic data into the downstream classification pipeline. The \textit{FairGen} framework ensures that both augmentation and loss reweighting are optimized to reduce residual biases.

\paragraph{Imbalance-aware Augmentation.}
The optimized diffusion model is used to generate high-quality synthetic samples for underrepresented $(r, d)$ pairs. By augmenting the dataset with these samples, we aim to balance the distribution of sensitive attributes and disease classes. Formally, the augmented dataset ${DS}_{\text{aug}}$ is defined as:
\begin{equation}
{DS}_{\text{aug}} = {DS} \cup \{\tilde{x}_i \mid \tilde{x}_i = \text{DM}(z_i; r, d), \forall i = 1, \dots, m_{r,d}\}
\label{eq:augmentation}
\end{equation}
where $\text{DM}$ represents the trained diffusion model and $z_i$ is the random noise input. This augmentation process (Eq.~\ref{eq:augmentation}) directly leverages the debiased synthetic data generated by the diffusion model, addressing both under-representation and data quality concerns.

\paragraph{Adaptive Inverse-Performance Reweighting.}
To rectify the decision boundary bias toward majority subgroups, we implement a dynamic, performance-aware reweighting mechanism during classifier training. Let $A_r^{(t)}$ denote the validation accuracy of the sensitive subgroup $r$ at epoch $t$. To prevent numerical instability when accuracy is low and to maintain consistent gradient magnitudes, we formulate the dynamic weight $w_r^{(t)}$ using a normalized inverse-performance objective:

\begin{equation}
w_r^{(t)} = K \cdot \frac{\left(A_r^{(t)} + \epsilon\right)^{-1}}{\sum_{j \in S} \left(A_j^{(t)} + \epsilon\right)^{-1}}
\label{eq:reweight}
\end{equation}

where S represents the set of all sensitive attributes, $\epsilon$ is a smoothing term to ensure numerical stability, and $K=|S|$ serves as a normalization factor to preserve the expected scale of the loss function. This strategy dynamically redistributes the optimization focus, penalizing the model more heavily for errors on underperforming subgroups.

\paragraph{Robust Evaluation Metrics.}
To validate the efficacy of the downstream balancing strategies, we employ fairness-aware evaluation metrics, such as Equal Opportunity (EO)\cite{hardt2016equality}, Demographic Parity (DP)\cite{dwork2012fairness}, Equalized Sensitive Subgroup Performance (ESSP)\cite{tian2023fairseg}, and Normalized Accuracy Range (NAR)\cite{groh2021evaluating} across all sensitive attributes and disease classes. These metrics quantify the reduction in bias and improvement in classification performance.

\subsection*{Clinician-Facing Expert Evaluation}

To complement automated image-quality metrics, we additionally conducted a blinded clinician-facing pairwise evaluation across the three modalities. Candidate images were drawn from original and generated pools and assembled into modality-specific comparison pairs under standardized disease-focused prompts. For each modality, 20 pairs were constructed for review, and the expert was asked to select which image better matched the pathology-relevant visual features specified by the prompt rather than which image merely appeared more realistic.

This protocol was designed to test whether pathology-relevant content remained recognizable in FairGen outputs. The resulting responses were summarized as pairwise preference rates across modalities and reported in the Results section. We emphasize that this evaluation was intended as a focused clinician-facing qualitative assessment rather than as a formal multi-rater agreement study; therefore, inter-rater reliability statistics such as ICC were not part of the current design.

\subsection*{Computational Resources and Reproducibility}

All major diffusion experiments were implemented with Hugging Face Diffusers and Accelerate. Training was performed on NVIDIA Tesla V100 32GB GPUs using fp16 mixed precision and gradient checkpointing. For the main diffusion fine-tuning runs, we used a per-device batch size of 8 with 4-step gradient accumulation, yielding an effective batch size of 32, together with a learning rate of \(3\times10^{-7}\), a cosine scheduler, and 15{,}000 optimization steps. In our current experiments, representative wall-clock training times ranged from approximately 18.7 to 32.7 hours per diffusion model depending on the method variant. We report these practical details to make the computational setup and reproduction burden of the experiments more transparent.

\section*{Data availability}

Due to privacy restrictions and licensing agreements associated with medical data, the specific preprocessed training subsets and the physician-aligned preference dataset generated in this study are not publicly available. However, de-identified data supporting the findings of this study may be available from the corresponding authors upon reasonable request.

\section*{Code availability}

The source code for the FairGen framework, including the implementation of the physician-aligned preference optimization and downstream evaluation pipelines, is publicly available on GitHub at \url{https://github.com/Krisocer/FairGen}.

\section*{Acknowledgements}
This work was partially funded by the National Institutes of Health (NIH) under grant number R01EB033387 and  1R01EB037101-
01. The funder had no role in the study design, analysis, interpretation of the results, or preparation of the manuscript. We acknowledge the University of Pittsburgh Medical Center (UPMC) for clinical guidance. We also acknowledge the creators and maintainers of the public datasets used in this study.

\section*{Author contributions}
Conceived and designed the experiments: Z.L., R.Z., Z.T, J.H., T.C. Performed the experiments: Z.L., R.Z., Z.T. Analyzed the data: Z.L., R.Z., Z.T., H.J.A., J.H., T.C. Wrote the paper: Z.L., R.Z., Z.T., H.J.A., J.H., T.C. All authors have read and agreed to the published version of the manuscript.

\section*{Competing interests}
The authors declare no competing financial or non-financial interests.

\bibliography{sample}

\clearpage
\section*{Figure legends}

\noindent\textbf{Figure 1. Overview of FairGen's pipeline for mitigating bias through generation and AI-based diagnosis.}
\textbf{(a)}, Data imbalance and task structure. Three panels highlight severe demographic disparities: long-tailed skin tone distributions and skewed age/gender ratios in MRI and radiography.
\textbf{(b)}, FairGen leverages diffusion-based synthesis with data balance, physician-aligned preference rewards, and class diversity to generate balanced images, reducing bias across underrepresented subgroups. (E: encoder transforms images into latent representations; D: decoder reconstructs images from latent representations.)
\textbf{(c)}, In the downstream stage, both real and generated data are combined to train classification models, with fairness improvements across sensitive attributes. This integrated workflow effectively addresses long-tail and skewed data problems, promoting fair performance for diverse underrepresented subpopulations.

\vspace{1em}
\noindent\textbf{Figure 2. Evaluations of synthetic images generated by FairGen compared to baseline methods.}
\textbf{(a)}, Quantitative image quality evaluation using Fr\'echet Inception Distance (FID; lower indicates better quality) comparing FairGen against baseline methods (SD, FairDiffusion, CBRS, SQRS, CBDM) across all modalities.
\textbf{(b)}, Percentage of relative quality improvement compared to the SD baseline, highlighting FairGen's significant enhancement across all modalities.
\textbf{(c)}, Visual comparisons demonstrating FairGen's improved clinical realism and anatomical detail relative to baseline methods and real clinical images.

\vspace{1em}
\noindent\textbf{Figure 3. Expert evaluation design and outcomes across three medical imaging modalities.}
\textbf{(a)}, Pair construction for expert review. Candidate images were drawn from original and generated image pools, and modality-specific comparison pairs were constructed for blinded review.
\textbf{(b)}, Questionnaire design. For each pair, experts were asked to select which image better matched disease-relevant clinical features under a modality-specific prompt.
\textbf{(c)}, Expert review outcomes. The left panel summarizes the distribution of expert choices across T1-weighted MRI, Skin Image, and Chest Radiography. The right panel summarizes the modality-level preference rates associated with the target comparison setting used in each expert review setup. Together, these results provide supportive clinician-facing evidence that FairGen preserves pathology-relevant visual features beyond standard fidelity metrics.

\vspace{1em}
\noindent\textbf{Figure 4. Multidimensional assessment of generative fidelity and diagnostic fairness across modalities.}
\textbf{(a)}, Downstream diagnostic fairness evaluation using radar charts. FairGen (yellow polygons) consistently outperforms SD (blue polygons) by achieving larger areas for ESSP (higher is better) and more compact, centralized shapes for bias metrics (DP, EO, NAR; lower is better).
\textbf{(b)}, Quantitative generative fairness assessment via Fr\'echet Inception Distance (FID). The left columns show FID distribution across demographic subgroups, while the right columns visualize the inter-subgroup FID variance.
\textbf{(c)}, Qualitative analysis of manifold alignment using t-SNE projections of latent features colored by sensitive attributes. While baseline methods (e.g., SD, CBRS) exhibit distinct clustering or segregation of subgroups (indicating learned bias), FairGen produces a homogenously mixed distribution.

\vspace{1em}
\noindent\textbf{Figure 5. Evaluation of diagnostic robustness across diverse classifier architectures and subgroup equity.}
\textbf{(a-d)}, Diagnostic Classification Accuracy Scores (CAS) under varying synthetic-to-real data ratios across four distinct machine learning architectures: \textbf{(a)} Vision Transformer (ViT-base-patch16-224), \textbf{(b)} K-Nearest Neighbors (KNN), \textbf{(c)} ResNet50, and \textbf{(d)} XGBoost. The downstream tasks are five-way Skin Image classification (Allergic Contact Dermatitis, Basal Cell Carcinoma, Lichen Planus, Psoriasis, Squamous Cell Carcinoma), five-way Chest Radiography classification (COVID-19, Edema, Lung Opacity, Pleural Effusion, Normal), and binary T1-weighted MRI classification (Nondemented vs.\ Demented). FairGen (solid orange line) demonstrates consistent robustness and superior performance compared to five baseline methods, proving that its utility is architecture-agnostic. In stable modalities (MRI, Chest Radiography), FairGen maintains or improves accuracy even at high augmentation ratios.
\textbf{(e)}, Subgroup-specific performance analysis comparing FairGen against the SD baseline using radar charts. These results are derived from the \textbf{Vision Transformer (ViT) model}, which was selected as the representative architecture due to its superior overall performance compared to other classifiers (KNN, ResNet, XGBoost). The polygons illustrate the Area Under the Curve (AUC) for each sensitive attribute subgroup. FairGen effectively closes the gap for underrepresented populations (e.g., improving dark skin tone and above-75 age-group accuracy) while maintaining high performance for majority groups, resulting in more symmetrical and fairer diagnostic profiles.

\vspace{1em}
\noindent\textbf{Figure 6. Evaluation of expert alignment, latent representations, and data characteristics.}
\textbf{(a)}, Physician preference annotation. Experts selected images based on anatomical fidelity (e.g., cortical atrophy) to serve as ground truth for Direct Preference Optimization (DPO).
\textbf{(b)}, UMAP projections of latent features, colored by disease labels, with silhouette scores reported above each panel. The visualization is used as a complementary view of embedding-space organization rather than a standalone performance metric.
\textbf{(c)}, Original dataset distributions. Pie charts quantify the severe imbalance across sensitive attributes (e.g., skin tone, age) and disease classes in the training data.

\end{document}